\documentclass[letterpaper, 10 pt, conference]{ieeeconf}

\IEEEoverridecommandlockouts                              

\usepackage{pgf}
\usepackage[linesnumbered, ruled]{algorithm2e}
\makeatletter
\newcommand{\nosemic}{\renewcommand{\@endalgocfline}{\relax}}
\newcommand{\dosemic}{\renewcommand{\@endalgocfline}{\algocf@endline}}
\let\oldnl\nl
\newcommand{\nonl}{\renewcommand{\nl}{\let\nl\oldnl}}
\makeatother
\usepackage{graphics} 
\usepackage{epsfig} 
\usepackage{times} 
\usepackage{amsmath} 
\usepackage{amssymb}  
\usepackage{psfrag}
\usepackage{booktabs}
\usepackage{siunitx}
\usepackage{svg}
\usepackage{pgfplots}
\usepackage{graphicx}
\usepackage{cite}

\usepackage{makecell}
\usepackage[utf8]{inputenc}
\DeclareUnicodeCharacter{2060}{\nolinebreak}
\usepackage{pgfplots}
\usepackage{pdfpages}

\usepackage{balance}
\usepgfplotslibrary{groupplots}

\usepackage{flexisym}

\SetKwInput{KwInput}{Input}                
\SetKwInput{KwOutput}{Output}
\SetKwInput{KwInitialize}{Initialize}
\usepackage{euscript}
\usepackage{stmaryrd}
\usepackage{float}
\usepackage{url}
\usepackage{subfigure}

\title{\LARGE \bf
Falsification-Based Robust Adversarial Reinforcement Learning
}

\author{Xiao Wang$^{1}$, Saasha Nair$^{1}$, and Matthias Althoff$^{1}$
\thanks{$^{1}$The authors are with the Department of Informatics, Technische Universität München, 85748 Garching, Germany.
        {\tt\small \newline  xiao.wang@tum.de, saasha.nair@tum.de, althoff@in.tum.de}}
}

\begin{document}

\maketitle
\thispagestyle{empty}
\pagestyle{empty}

\begin{abstract}

Reinforcement learning (RL) has achieved enormous progress in solving various sequential decision-making problems, such as control tasks in robotics. Since policies are overfitted to training environments, RL methods have often failed to be generalized to safety-critical test scenarios. Robust adversarial RL (RARL) was previously proposed to train an adversarial network that applies disturbances to a system, which improves the robustness in test scenarios. However, an issue of neural network-based adversaries is that integrating system requirements without handcrafting sophisticated reward signals are difficult. Safety falsification methods allow one to find a set of initial conditions and an input sequence, such that the system violates a given property formulated in temporal logic. In this paper, we propose falsification-based RARL (FRARL): this is the first generic framework for integrating temporal logic falsification in adversarial learning to improve policy robustness. By applying our falsification method, we do not need to construct an extra reward function for the adversary. Moreover, we evaluate our approach on a braking assistance system  and an adaptive cruise control system of autonomous vehicles. Our experimental results demonstrate that policies trained with a falsification-based adversary generalize better and show less violation of the safety specification in test scenarios than those trained without an adversary or with an adversarial network.

\end{abstract}

\section{Introduction}
\label{sec:intro}

Recent advancements, such as superhuman performance in a range of Atari games in 2015 \cite{mnih2015human}, followed by AlphaGo’s victory against the human world champion in Go in 2016 \cite{silver2016mastering, chouard2016go}, have developed numerous research interests in reinforcement learning (RL) \cite{arulkumaran2017brief}. Consequently, RL has greatly progressed in real-world applications, such as robotics \cite{polydoros2017survey}, natural language processing \cite{keneshloo2019deep}, and autonomous driving \cite{kendall2019learning, Krasowski2020a}. However, RL still suffers from some shortcomings, such as bad generalization in real-world scenarios, risk-sensitive reward functions, and violation of safety constraints \cite{dulac2019challenges, rlblogpost}. This study addresses the generalization problem due to the huge amount of training required for RL.

Pinto et al. \cite{pinto2017robust} discussed generalization by adding disturbances as adversarial examples \cite{goodfellow2014explaining}. Their study was later extended by Pan et al. \cite{pan2019risk}. By training with adversarial examples, the authors reduced the simulation-to-reality gap caused by modeling errors, so the trained models generalize better in real-world scenarios. Adversarial RL is formulated as a two-player zero-sum game in \cite{pinto2017robust, pan2019risk}, in which an adversary aims to obstruct the success of the learning system. However, learning in a zero-sum game requires finding a Nash equilibrium, which is especially challenging for continuous high-dimensional problems \cite{salimans2016improved}. Otherwise, if we formulate the problem as a non-zero-sum game, in which the adversary optimizes a different reward function, then a sophisticated reward function for the adversary would have to be handcrafted. 
One can argue that engineering the reward function could improve the generalization ability of an RL agent. However, as pointed out in \cite{rlblogpost}, designing a \emph{perfect} reward function is a very challenging task.
For instance, the traffic rule \emph{a vehicle is not allowed to overtake another on its right side except in congested traffic} requires a sequence of events, which is difficult to integrate into a reward function; the traffic rule can easily be expressed by a temporal logic. Therefore, temporal logic falsification methods provide a possibility to automatically improve generalization without having to tune the reward functions.

In this paper, we propose a new framework: we develop adversarial samples in a single RL agent setting, wherein the protagonist is represented by an RL agent, while safety falsification methods act as an adversary. Safety falsification approaches drive a system to unsafe behaviors, which violate given safety specifications \cite{abbas2013probabilistic}.

The remainder of this paper is organized as follows. Section~\ref{sec:related} provides an overview of current solutions in related studies for adversarial RL and system falsification for safety-critical systems. Section~\ref{sec:approach} introduces falsification-based RARL, which is evaluated in Section~\ref{sec:experiments}. In Section~\ref{sec:conclusions}, we present the conclusions and potential future research directions.





\section{RELATED WORK}
\label{sec:related}

\subsection{Adversarial Reinforcement Learning}

Despite the success of RL algorithms, they are susceptible to changes in environmental settings \cite{rlblogpost, morimoto2005robust}. Hence, various forms of adversarial training \cite{goodfellow2014explaining} have been introduced to solve this problem. One such approach involves adding adversarial perturbations to the observations of the agent by attacking either only the image inputs \cite{huang2017adversarial, lin2017tactics, kos2017delving} or addressing the entire state vector \cite{mandlekar2017adversarially, pattanaik2018robust}. Another approach involves using source-domain ensembles, which are adapted to the target domain using the Bayesian model adaptation \cite{rajeswaran2016epopt}. 

Further, the approach most relevant to our study is the minimax approach extending robust RL \cite{morimoto2005robust}. This approach, which is known as robust adversarial RL (RARL) \cite{pinto2017robust}, simultaneously trains two RL agents: one called the protagonist, while the other is called the adversary. The adversary is tasked with applying destabilizing forces to impede the protagonist, while the protagonist learns to be robust to the adversary. 
An extension of RARL has been provided by risk-averse RARL (RARARL) \cite{pan2019risk}, which focuses on safety-critical cyber-physical systems, by modeling the risk as the variance of an ensemble of value functions. However, RARARL can only solve problems with a discrete action space and requires an ensemble of multiple neural networks, requiring a significant computational resource. 
In contrast to the RARL and RARARL that model the setup as a two-agent RL scenario, as mentioned in Section~\ref{sec:intro}, our proposed solution reduces the number of reward functions to be defined and tuned, requires fewer parameters of the adversary to be optimized, as introduced in Section~\ref{subsec:safe}, and allows for better expressiveness for the adversary using temporal logic specifications.

\subsection{Safety Falsification}
\label{subsec:safety}

Safety falsification methods aim at finding initial conditions and input sequences, with which a system violates a given safety specification. Two categories of various approaches exist for solving this problem. We first review \emph{single-shooting} methods, which simulate trajectories from specific initial conditions and input traces and iterate until a falsifying trajectory is obtained. A single-shooting method is achieved by applying Monte-Carlo methods \cite{abbas2011linear, abbas2013probabilistic, nghiem2010monte}, ant colony optimization method \cite{annapureddy2010ant}, cross-entropy method \cite{sankaranarayanan2012falsification}, or rapidly-exploring random tree search \cite{bhatia2004incremental, dreossi2015efficient, dreossi2017compositional, koschi2019computationally}. A \emph{multiple-shooting} approach is proposed in \cite{zutshi2013trajectory, zutshi2014multiple} to split system trajectories into small segments by simulating from multiple initial conditions in a state space decomposed into cells. Once a segment reaches an unsafe state, the cell size is refined until the segments can be concatenated to a complete system trajectory. The ant-colony method and the Monte-Carlo method were compared in \cite{annapureddy2010ant} for two benchmarks, and similar results were obtained. As shown in \cite{sankaranarayanan2012falsification}, the cross-entropy method outperformed the Monte-Carlo method on five benchmarks. Hence, we employ the cross-entropy method in this study. Note that our framework can also use other falsification approach.

%
%
%
%
%
%
%




\section{FALSIFICATION-BASED RARL}
\label{sec:approach}

\subsection{Safety Falsification} 
\label{subsec:safe}

To formulate the safety falsification problem, we first introduce several important definitions adopted from \cite{abbas2013probabilistic}. A dynamic system $\Sigma$ can be regarded as a mapping from initial states $\boldsymbol{x}_0 \in \mathcal{X}_0 \subset \mathbb{R}^n $ and input signals $\boldsymbol{u} \in \mathcal{U} \subset \mathbb{R}^m$ to output signals $\boldsymbol{y} \in \mathcal{Y} \subset \mathbb{R}^k$: $\mathcal{X}_0 \times \mathcal{U} \rightarrow \mathcal{Y}$. 

We formulate system properties in metric temporal logic (MTL) \cite{koymans1990specifying}. A temporal logic combines propositions of classical logics with time dependence such that a truth value is assigned to each atomic proposition at each time instant \cite{temporallogic}. An atomic proposition $p$ is a statement that can be either \emph{true} or \emph{false}. Atomic propositions and the logical connectives, such as Boolean operators $not$ and $or$ denoted by $\neg$ and $\lor$, form propositional formulas. The temporal operator \emph{until}, which is denoted by $\boldsymbol{U}$, indicates that in a formula $\varphi_1 \, \boldsymbol{U} \, \varphi_2$, the first formula $\varphi_1$ holds \emph{until} the second formula $\varphi_2$ holds; the time $t$ when $\varphi_2$ starts to hold is unconstrained, i.e., $t\in (0, \infty)$. The temporal operator \emph{globally}, which is denoted by $\boldsymbol{G}$, indicates that formula $\varphi$ must hold for all times. In addition, MTL is an extension of temporal logic in which temporal operators are replaced by time-constrained operators. Thus, $\boldsymbol{U}$ is replaced by $\boldsymbol{U_I}$, where $\boldsymbol{I}\subseteq (0, \infty)$, indicating that $t$ is constrained by $\boldsymbol{I}$. The syntax of an MTL formula $\varphi$ is defined as follows \cite{nghiem2010monte}:
\begin{equation}
\label{eq:MTL_syntax}
	\varphi := true \, | \, p \, | \, \neg \varphi \, | \, \varphi_1 \lor \varphi_2 \, | \, \varphi_1 \, \boldsymbol{U_I} \, \varphi_2 | \boldsymbol{G} \varphi,
\end{equation}
which indicates that the value of an MTL formula is always \emph{true} or \emph{false}. Other logical expressions can be formed from logical equivalences, such as $a \implies b \equiv \lnot a \lor b$. In this study, we use the \emph{global} operator $\boldsymbol{G}$ as presented in \eqref{eq:MTL_acc}. More general formulas can be obtained from \eqref{eq:MTL_syntax}.

\textbf{\emph{Definition.}} (MTL Falsification). For an MTL specification $\varphi$, the MTL falsification problem aims to find initial states $\boldsymbol{x}_0 \in \mathcal{X}_0$ and an input sequence $\boldsymbol{u}: [0, T] \rightarrow \mathcal{U}$ such that the resulting trajectory $\boldsymbol{y}$ of system $\Sigma$ violates the specification $\varphi$, which is denoted by
\begin{equation}
\label{eq:falsification}
	\boldsymbol{y}(\boldsymbol{x}_0, \boldsymbol{u}) \not \models \varphi.
\end{equation}

Na\"ive falsification uniformly samples the set of initial conditions and input sequences. A more efficient approach is to guide the search using a metric, measuring the distance between the trajectory and set of states violating the specification. A \emph{robustness metric} $\varepsilon$ is proposed in \cite{fainekos2009robustness} to express the satisfaction of an MTL property over a given trajectory as a real number instead of a Boolean value ($0$ for no intersection with unsafe sets and $1$ for successful falsification). The sign of $\varepsilon$ reveals whether a trajectory $\boldsymbol{y}$ satisfies an MTL property $\varphi$. The robustness of $\boldsymbol{y}$ with respect to $\varphi$ is denoted by
\begin{equation}
 \label{eq:robustness}
 \varepsilon = \llbracket \varphi \rrbracket_d (\boldsymbol{y}, t),
\end{equation}
%
%
and defined as follows \cite{nghiem2010monte, sankaranarayanan2012falsification, annapureddy2010ant}:
\begin{equation}
\label{eq:MTLrobustness}
\begin{aligned}
	\llbracket true \rrbracket_d (\boldsymbol{y}, t) &:= +\infty \\
	\llbracket p \rrbracket_d (\boldsymbol{y}, t) &:= \mathbf{Dist}_d ( \boldsymbol{y}(t), \mathcal{O}(p) ) \\
	\llbracket \neg \varphi \rrbracket_d (\boldsymbol{y}, t) &:= - \llbracket \varphi \rrbracket_d (\boldsymbol{y}, t) \\
	\llbracket \varphi_1 \lor \varphi_2 \rrbracket_d (\boldsymbol{y}, t) &:=
		\max (\llbracket \varphi_1 \rrbracket_d (\boldsymbol{y}, t), \llbracket \varphi_2 \rrbracket_d (\boldsymbol{y}, t)) \\
	\llbracket \varphi_1 \, \boldsymbol{U_I} \, \varphi_2 \rrbracket_d (\boldsymbol{y}, t) &:= 
		\sup_{t' \in (t + \boldsymbol{I})} \min ( \llbracket \varphi_2 \rrbracket_d (\boldsymbol{y}, t'), \\
		&\qquad \inf_{t < t'' < t'} \llbracket \varphi_1 \rrbracket_d (\boldsymbol{y}, t'')),
\end{aligned}
\end{equation}
where $\mathcal{O}(p)$ denotes the set in which $p$ is fulfilled. The signed distance denoted by $\mathbf{Dist}_d$ is defined as
\begin{equation}
\label{eq:signedDist}
	\mathbf{Dist}_d ( \boldsymbol{y}, \mathcal{O} ) :=
	\left\{
  \begin{array}{lr}
    -\inf \{ d(y,x) \, | \, x \in \mathcal{O} \} & \mathrm{if} \, y \notin \mathcal{O} \\
    \inf \{ d(y,x) \, | \, x \in \mathcal{X} \setminus \mathcal{O} \} & \mathrm{if} \, y \in \mathcal{O} ,
  \end{array}
	\right.
\end{equation}
where $d(y,x)$ is typically defined as the Euclidean distance for continuous systems: 
\begin{equation}
 \label{eq:state}
 d(y, x) = || y - x ||^2.
\end{equation}
Consequently, \eqref{eq:falsification} can be defined as a minimization problem:
\begin{equation}
 \label{eq:minrobustness}
 \min_{\boldsymbol{x}_0 \in \mathcal{X}_0, \boldsymbol{u} : [0,T] \to \mathcal{U}}
		\llbracket \varphi\rrbracket_d(\boldsymbol{y}(\boldsymbol{x}_0, \boldsymbol{u}), t) .
\end{equation}

The cross-entropy method combines piecewise-uniform and Gaussian distributions to approximate the underlying distribution of the robustness value in \eqref{eq:robustness} over the set $\mathcal{X}_0 \times \mathcal{U}$ \cite{sankaranarayanan2012falsification}. The proposed distribution is denoted by $p_{\theta}$ with parameter $\theta$, while the unknown real distribution is denoted by $q$. The distance between the two distributions is measured using the Kullback-Leibler divergence \cite{kullback1951information}:
\begin{equation}
\label{eq:kldivergence}
D(q,\, p_\theta) = \int_{\mathcal{X}_0 \times \mathcal{U}} \log \left( \frac{q(\xi)}{p_\theta(\xi)} \right) q(\xi) \, \mathrm{d}\xi,
\end{equation}
where $\xi \in \mathcal{X}_0 \times \mathcal{U}$. Since the actual distribution $q$ is unknown, $D(q, \,p_\theta)$ is estimated using $N_s$ sampled data points, which are chosen by the current approximation $p_\theta$. The samples are sorted through their robustness values, and the $m$ least robust samples are considered, with $m\ll N_s$. Then, parameter $\theta$ is updated by minimizing the divergence $D(q, \,p_\theta)$ over $m$ data samples. Moreover, this procedure iterates until the divergence converges to a threshold. Afterward, the initial conditions and input sequences are sampled based on the converged distribution $p_\theta$.

In this study, we consider an autonomous vehicle on a highway with two safety requirements: the agent is not allowed to collide with the leading vehicle or to drive backward on the highway at all times. Therefore, we formulate these requirements as follows:
\begin{equation}
\label{eq:MTL_acc}
\boldsymbol{G} (\lnot \varphi_{\mathrm{collision}} \wedge \lnot \varphi_{\mathrm{reverse}}).
\end{equation}
%
Note that the proposed method can be directly applied to more complicated specifications containing temporal operators, such as \emph{until} and \emph{eventually}. In the future, we will integrate more traffic rules in the system requirements as proposed in \cite{rizaldi2015formalising, rizaldi2017formalising, esterle2019specifications, esterle2020formalizing, Maierhofer2020a}.

\subsection{Falsification-Based RARL}
\label{subsec:framework}

\begin{figure}[tb]
\centering
\footnotesize
\makebox[0pt]{\includegraphics[scale=0.45]{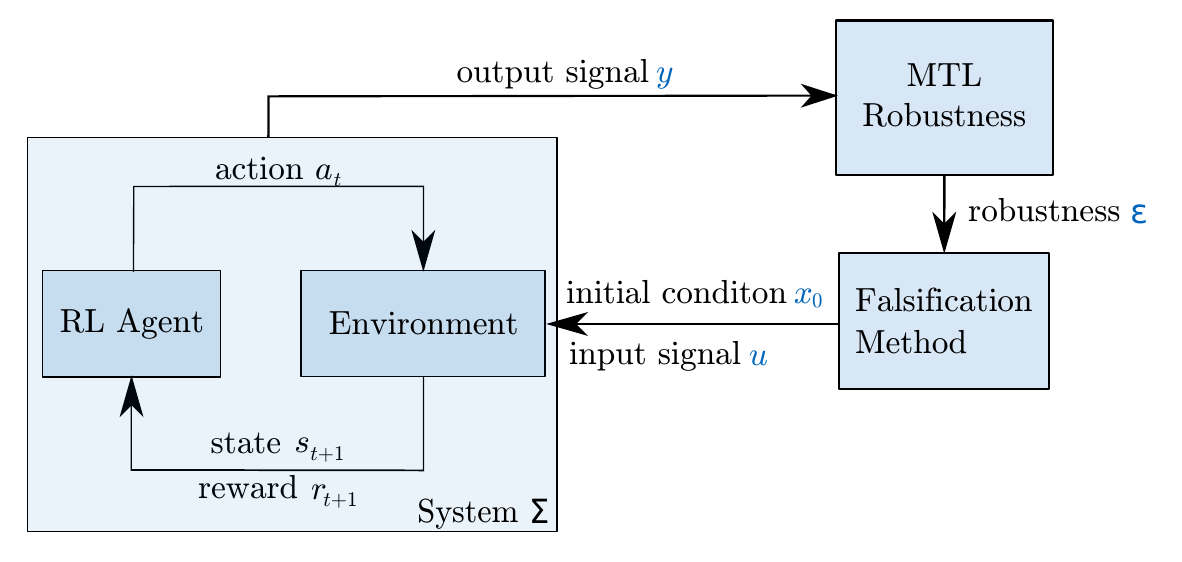}}
\caption{Falsification-based RARL framework. The RL agent and environment are regarded as the black box system $\Sigma$. Falsification model serves as the adversary, which provides input sets $x_0$ and $u$ for the environment such that the system trajectory falsifies MTL specifications. The RL agent is trained further under the falsified environment.}
\label{fig:framework}
\end{figure}

Figure~\ref{fig:framework} depicts our framework of falsification-based RARL, and Algorithm~\ref{algo:FRARL} presents our approach in detail. We formulate our RL problem as a Markov decision process (MDP) defined by a 5-tuple $(S, A, P, R, \gamma)$, where $S$ denotes the state space, $A$ denotes the action space, $P$ denotes the state transition probability, $R$ denotes the expected reward signal, and $\gamma \in [0, 1]$ denotes the discount factor. Further, the left part of Fig.~\ref{fig:framework} describes the learning process of the RL policy. 
The agent acts on the environment, followed by updating its state and reward. Our experiments reveal that the policy converges slower if the adversary interferes too early, as was also observed in \cite{pan2019risk}. Therefore, as shown in Algorithm~\ref{algo:FRARL} line~\ref{alg:ppo_start}-\ref{alg:ppo_end}, we first train the agent for $t_f$ time steps without an adversary (see Fig.~\ref{fig:learning_curve_highd} and Fig.~\ref{fig:learning_curve_random}). Next, we regard our trained model and the environment as a black box system $\Sigma$. As illustrated in the right part of Fig.~\ref{fig:framework}, we employ the cross-entropy method to obtain initial conditions and input sequences for the environment (the behavior of other vehicles) under which the agent violates our MTL specification \eqref{eq:MTL_acc}. 
We initialize our environment with the initial conditions, change the behavior based on the new input sequences, and train the policy further in the new adversarial environment (line~\ref{alg:env}). 
This procedure is repeated until the policy converges to zero violations. 

%
\begin{algorithm}
\label{algo:FRARL}
\SetAlgoLined
\caption{Falsification-Based RARL}
  \KwInput{Training steps $T$; environment $E$; number of actors $n_a$; time steps each actor runs at each iteration $t_a$; MTL specification $\varphi$; time step to start falsification $t_f$; number of falsification iterations $n_f$}
  \KwInitialize{Parameters of policy and value network $\phi_0$}
  \KwResult{Trained policy and value network $\phi$}
  \While{$t < T$}{
        \For{$\mathrm{actor}=1,2,...,n_a$} 
        { \label{alg:ppo_start}
        Run policy $\phi_{\mathrm{old}}$ in $E$ for $t_a$ time steps \;
        Compute advantage estimates $\hat{A}_1, ..., \hat{A}_{t_a}$ \eqref{eq:advantage}\;
        }        
        Optimize surrogate $L^{\mathrm{CLIP+VF}}$ \eqref{eq:ppo} wrt. $\phi$ with batch size $n_a t_a$ \;
        $\phi_{\mathrm{old}} \leftarrow \phi$ \;
        $t = t + n_at_a$ \; \label{alg:ppo_end}
        \If{$t > t_f$}
        {
        Initialize falsifier parameter $\theta_0$ \;
            \For{$\mathrm{iter}=1,2,...,n_f$}
            {
            Sample input conditions $\boldsymbol{x}_0, \boldsymbol{u}$ from $p_{\theta_{\mathrm{old}}}$ \; 
            Initialize new environment $E_{\mathrm{iter}}$ with $\boldsymbol{x}_0$, change the behavior of $E_{\mathrm{iter}}$ with $\boldsymbol{u}$ \; \label{alg:env}
            Collect trajectories $\boldsymbol{y}$ in $E_{\mathrm{iter}}$ with agent $\phi_{\mathrm{old}}$ and evaluate robustness value according to \eqref{eq:robustness} \;
            Estimate \eqref{eq:kldivergence} with $\boldsymbol{y}$ and minimize \eqref{eq:kldivergence} wrt. $\theta$ \;
            $\theta_{\mathrm{old}} \leftarrow \theta$ \;
            }
            $E \leftarrow E(\boldsymbol{x}_0, \boldsymbol{u}) $ \;
        }
    }
\end{algorithm}

We choose proximal policy optimization (PPO) \cite{schulman2017proximal} to optimize the policy network due to its superior performance in continuous control problems when compared with other state-of-the-art approaches. To reduce variance, we use an actor-critic architecture \cite{konda2000actor} to approximate both the policy and the value function with neural networks. Additionally, we estimate the advantage function $\hat{A}_t$ using a general advantage estimator (GAE) \cite{schulman2015high} as follows:
\begin{equation}
\begin{aligned}
\label{eq:advantage}
&\hat{A}_t = \delta_t + (\gamma \lambda)\delta_{t+1}+...+...+(\gamma \lambda)^{T-t+1}\delta_{T-1}, \\
&\mathrm{with} \quad  \delta_t = r_t + \gamma V(s_{t+1}) - V(s_t),
\end{aligned}
\end{equation}
where $\lambda \in [0, 1]$ denotes a discount factor of the advantage estimator and makes a compromise between variance and bias. The objective function of PPO is defined as follows:
\begin{subequations}\label{eq:ppo}
\hspace*{-0.2cm}\vbox{
\begin{align}
&L^{\mathrm{CLIP+VF}}(\phi) =\hat{\mathbb{E}}_t \left[L_t^{\mathrm{CLIP}}(\phi) - c L_t^{\mathrm{VF}}(\phi) \right ], \mathrm{with} \\
&L^{\mathrm{CLIP}}(\phi)    =\hat{\mathbb{E}}_t \left[\min(r_t(\phi)\hat{A}_t, \mathrm{clip}(r_t(\phi), 1-\epsilon, 1+\epsilon)\hat{A}_t)  \right] \\
&L_t^{\mathrm{VF}}(\phi)    = (V_\phi(s_t) - V_t^{\mathrm{targ}})^2,
\end{align}}
\end{subequations}
where $\phi$ denotes the parameters of the policy and value network, the probability ratio is defined by $r_t(\phi)= \dfrac{\pi_\phi(a_t|s_t)}{\pi_{\phi_{old}}(a_t|s_t) }$, $\pi_{\phi_{old}}$ denotes the old policy before the update, $c$ and $\epsilon$ are hyper-parameters, $V_\phi$ denotes the estimated value function, $V_t^{\mathrm{targ}}$ is the target value function collected through Monte-Carlo simulations, $\mathrm{clip}$ is an operator for limiting the operand in a given range, and $\hat{\mathbb{E}}_t[...]$ denotes the empirical average over a finite batch of samples.  

\section{EXPERIMENTS}
\label{sec:experiments}

We evaluate our approach on two systems. The first one is a braking assistance (BA) system of an autonomous vehicle applied to avoid rear-end collisions and driving reversely on a highway, while the second one is an adaptive cruise control (ACC) system that keeps a safe distance from the leading vehicle and follows the desired velocity. 
The two systems are implemented in the same traffic simulator with different reward functions, which are described in Section~\ref{subsec:env}. To fairly evaluate the performance of our approach, we train each system in three environments: a baseline environment without an adversarial model, an adversarial environment with an RL agent as an adversary, and an adversarial environment with an adversary using our falsification method. The adversarial RL agent in the second environment is trained by utilizing RARL \cite{pinto2017robust}. Moreover, because RARARL \cite{pan2019risk} was proposed to solve discrete action space problem and cannot be directly applied to problems with a continuous action space, we choose RARL over the more recent RARARL to train the adversarial RL agent. Following \cite{pinto2017robust}, we call the policy that controls the ego vehicle \emph{protagonist}.

\subsection{Dataset}

In the adversarial environments, the behavior of the leading vehicle is altered by applying the falsification method or an adversarial RL agent. The baseline environment could be achieved by utilizing either rule-based driver models, such as the intelligent driver model (IDM) \cite{treiber2000congested}, or real traffic data. A key limitation of rule-based driver models is their homogeneity. The policy could easily overfit to react only to a particular behavior such that it fails to generalize, while driving behaviors from real traffic are more diverse. Therefore, we choose the recently published highway drone (HighD) dataset of naturalistic vehicle trajectories on German highways \cite{krajewski2018highd}. 

\begin{figure}[tb]
\centering
\footnotesize
\vspace{0.2cm}{
\makebox[0pt]{\includegraphics[trim={0 0 0 0.5cm}, clip, scale=0.53]{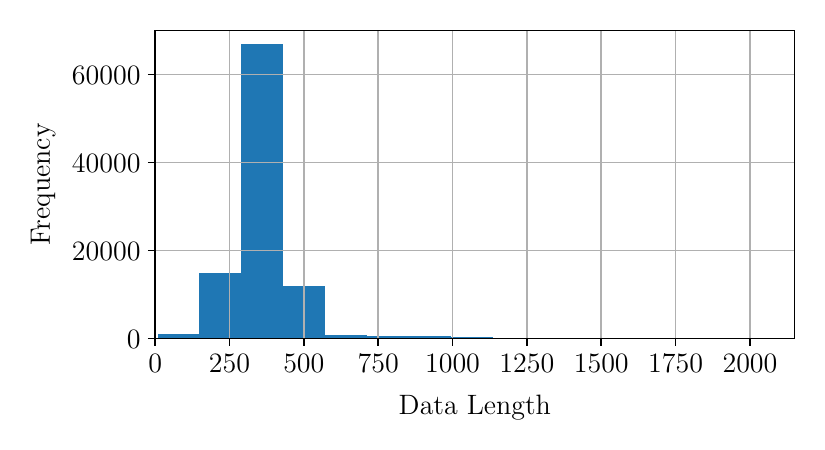}}}
\caption{Frequency histogram of the length of lane-following trajectories in HighD dataset \cite{krajewski2018highd}.}
\label{fig:length}
\end{figure}

HighD recorded \SI{16.5}{h} of video at six locations using a drone and it extracted over \SI{45000}{km} of vehicle trajectories at \SI{25}{Hz} using computer vision algorithms. Since the longitudinal driving behavior of a lane-changing vehicle differs from that of a lane-following vehicle, we filter out the trajectories of all lane-changing vehicles so that \SI{97184}{} lane-following trajectories remain. 
As depicted in Fig.~\ref{fig:length}, the frequency histogram demonstrated the distribution of the total length of these trajectories. To avoid overfitting, each trajectory should be used at most once during training. In addition, the original traffic scenarios cover a lane of \SI{420}{m} with a median duration of \SI{13.6}{s} for each vehicle. In this setting, the ego vehicle is less likely to encounter a critical situation. Therefore, we extend the lane length to \SI{600}{m} and the total time of a scenario to \SI{20}{s}, i.e., \SI{500}{} time steps. Therefore, to obtain sufficient trajectories, we select the longitudinal acceleration signals of lane-following trajectories with total time step $L\geq250$, cut signals to \SI{250}{} time steps, and append the signals with reversely duplicated signals. In all, we obtain \SI{93454}{} trajectories and separate them into \SI{70}{\%} and \SI{30}{\%} trajectories for training and testing, respectively.

\subsection{Environment}
\label{subsec:env}

We set up a driving simulator based on the CommonRoad benchmark suite \cite{commonroad} and OpenAI Gym \cite{openaigym}. Since the goal of the agent is to learn the longitudinal driving behavior, our simulator contains only a straight lane of \SI{600}{m}. An episode terminates if the leading vehicle reaches the end of the lane, the maximal time step \SI{500}{} is reached, a collision happens, or the ego vehicle drives in reverse. Both vehicles are driven based on a point-mass model, whose input is
acceleration, which is sampled from the policy network. To ensure that the scenario is solvable, the leading vehicle is initially at least as far away from the ego vehicle as the safe distance. We assume that both vehicles have the same maximum deceleration $a_{\mathrm{max}}=\SI{10}{m/s^2}$. Then, the safe distance is computed based on \cite{safedistance} as follows:
\begin{equation}\label{eq:safe_distance}
s_{\mathrm{safe}} = \frac{1}{2a_{\mathrm{max}}}(v^2_{f} - v^2_{l}) + v_{f} \delta,
\end{equation}
where $v_{f}$ and $v_{l}$ denote the velocities of the following and leading vehicles, respectively, and $\delta$ is the reaction delay of the following vehicle. As assumed by \cite{safedistance}, we utilize $\delta=0.3\,\si{s}$ for autonomous vehicles. 

Without loss of generality, the initial position of the ego vehicle is fixed at $s_{\mathrm{ego}}=\SI{10}{m}$, whereas the initial position of the leading vehicle is either randomly sampled within the range $\left[ s_{\mathrm{ego}}+s_{\mathrm{safe}},\, s_{\mathrm{ego}} + s_{\mathrm{safe}} +40\right]$ or calculated by applying the falsification tool. The acceleration and initial velocity of the leading vehicle are either extracted from the selected HighD trajectories, or computed by utilizing the adversaries, e.g., in \eqref{eq:minrobustness}, $\boldsymbol{x}_0$ corresponds to the initial position and velocity of the leading vehicle, and $\boldsymbol{u}$ corresponds to the acceleration of the leading vehicle.

Since maintaining a safe distance from the leading vehicle is crucial for avoiding collision, the feature vector of the policy networks should provide all the necessary information to calculate the safe distance in \eqref{eq:safe_distance}. Thus, we choose the feature vector for both systems as presented in Tab.~\ref{tab:feature}.

The reward function of the protagonist of the BA system is defined as follows:
\begin{equation}\label{eq:reward}
 r_{\mathrm{BA}} =
 \begin{cases}
  -1,& \text{if agent drives reversely or collision happens} \\
  0, & \text{otherwise}.
 \end{cases}
\end{equation}
The reward function of the protagonist of the ACC system is defined as follows:
\begin{equation}
\label{eq:continuous_reward}
r_{\mathrm{ACC}} =
\begin{cases}
&-1,\text{if agent drives reversely or collision happens} \\
&-0.1 \exp{\left(\frac{-5\,s}{s_{\mathrm{safe}}}\right)}, \text{if $s < s_{\mathrm{safe}}$}\\
&-0.05 \exp{\left(\frac{-5\,v_{\mathrm{ego}}}{v_{\mathrm{leading}}}\right)}, \text{if $v_{\mathrm{ego}} < v_{\mathrm{leading}}$}\\
&0, \text{otherwise}.
\end{cases}
\end{equation}
The first and last items are the same as in $r_{\mathrm{BA}}$. Additionally, two terms are added: the second term penalizes a violation of the safe distance using a nonlinear function that increases the penalization as the ego vehicle gets closer to the leading vehicle; the third term penalizes the ego vehicle for driving slower than the leading vehicle if the distance is greater than the safe distance. Note that the coefficients in \eqref{eq:reward} and \eqref{eq:continuous_reward} are selected through a grid search. The nonlinear functions in the second and third terms in \eqref{eq:continuous_reward} significantly increase the performance of the agent.

The goal of the adversarial policy is to minimize the reward of the protagonist. Therefore, we choose $r_{\mathrm{adv}} = -r_{\mathrm{BA}}$ and $r_{\mathrm{adv}} = -r_{\mathrm{ACC}}$ as the reward functions of the adversarial policies for the BA and ACC systems.

\begin{table}[tb]
\caption{Features used by policy network}
\label{tab:feature}
\vspace{-0.3cm}\vbox{\begin{center}
\begin{tabular}{l l l}
\toprule[0.5pt]
Feature & Units & Description\\
\midrule[0.25pt]
$s$                                           & \si{m}       & distance to leading vehicle \\
$v_{\mathrm{ego}} - v_{\mathrm{leading}}$     & \si{m/s}     & relative velocity to leading vehicle \\
$v_{\mathrm{ego}}$                            & \si{m/s}     & velocity of ego vehicle \\
$a_{\mathrm{leading}}$                        & \si{m/s^2}   & acceleration of leading vehicle \\
$a_{\mathrm{ego}}$                            & \si{m/s^2}   & acceleration of ego vehicle \\
\bottomrule[0.5pt]
\end{tabular}
\end{center}}
\vspace{-0.2cm}
\end{table}


\subsection{Baseline Model}
\label{subsec:baseline} 

As mentioned in Section~\ref{subsec:framework}, we train our policies and value functions by utilizing PPO \cite{schulman2017proximal} and an actor-critic algorithm \cite{konda2000actor}.
In particular, we use a shared network design to share the features between the policy and the value functions. Moreover, we build our models based on the implementation of OpenAI Baselines \cite{baselines}. Since our goal is to compare all methods with the same hyper-parameters, we did not perform a hyper-parameter optimization for each method. We instead utilized the default hyper-parameters that the OpenAI Baselines \cite{baselines} provides. The shared policy and value network has two hidden layers with 64 neurons each and \emph{tanh} as its activation function. The model is optimized using Adam optimizer \cite{kingma2014adam} with a learning rate of 0.0003 and a batch size of 128. In \eqref{eq:advantage}, the discount factor of the advantage estimator is $\lambda=0.95$.

In all conducted experiments, we first train the protagonist policy without the adversary for 200,000 training steps to allow it to learn basic skills as proposed in \cite{pan2019risk}. In the RL adversarial environment, we update the parameters of the protagonist $\theta_{\mu}$ to maximize $r_{\mathrm{BA}}$ or $r_{\mathrm{ACC}}$ for $N_\mu=10$ iterations, while the parameters of the adversary $\theta_{\nu}$ are kept constant. Then, to maximize $r_{\mathrm{adv}}$, we keep $\theta_{\mu}$ constant and update $\theta_{\nu}$ for $N_\nu=1$ iteration. Here, $N_\mu$ and $N_\nu$ are empirically chosen. This process iterates until both policies converge. In the falsification adversarial environment, we apply S-Taliro \cite{annpureddy2011staliro}, which is a MATLAB toolbox for MTL falsification for hybrid systems, to falsify the protagonist during training. S-Taliro is called every $10$ iterations to compute $10$ acceleration traces as well as initial positions and velocities for the leading vehicle, in which the protagonist falsifies the given specification \eqref{eq:MTL_acc}. To train the policy further, the computed traces were randomly picked by the simulator. For the rest of this paper, we call the baseline model PPO, the policy model trained with an RL adversarial agent RARL, and the policy model trained with our method FRARL.

\subsection{Evaluation}

During the training phase, policies are set to be stochastic to encourage exploration, whereas during the evaluation and falsification phases, deterministic policies are used. To fairly compare the robustness of the three models, $10$ policies with different random seeds are trained for each method. We evaluate the learning progress of all models in two groups of test scenarios, namely the HighD and random test scenarios, where the acceleration of the leading vehicle is randomly sampled in a given range. Note that we used random test scenarios instead of the adversarial or falsified scenarios because the agent is destined to encounter low reward in the adversarial and falsified scenarios.

\begin{figure*}[tb]
	\centering
	\footnotesize
	\subfigure[Episode reward and safe distance violation curves on the HighD testing scenarios of the braking assistant system.]{\includegraphics[trim={0 0 0 0.3cm}, clip, scale=0.375]{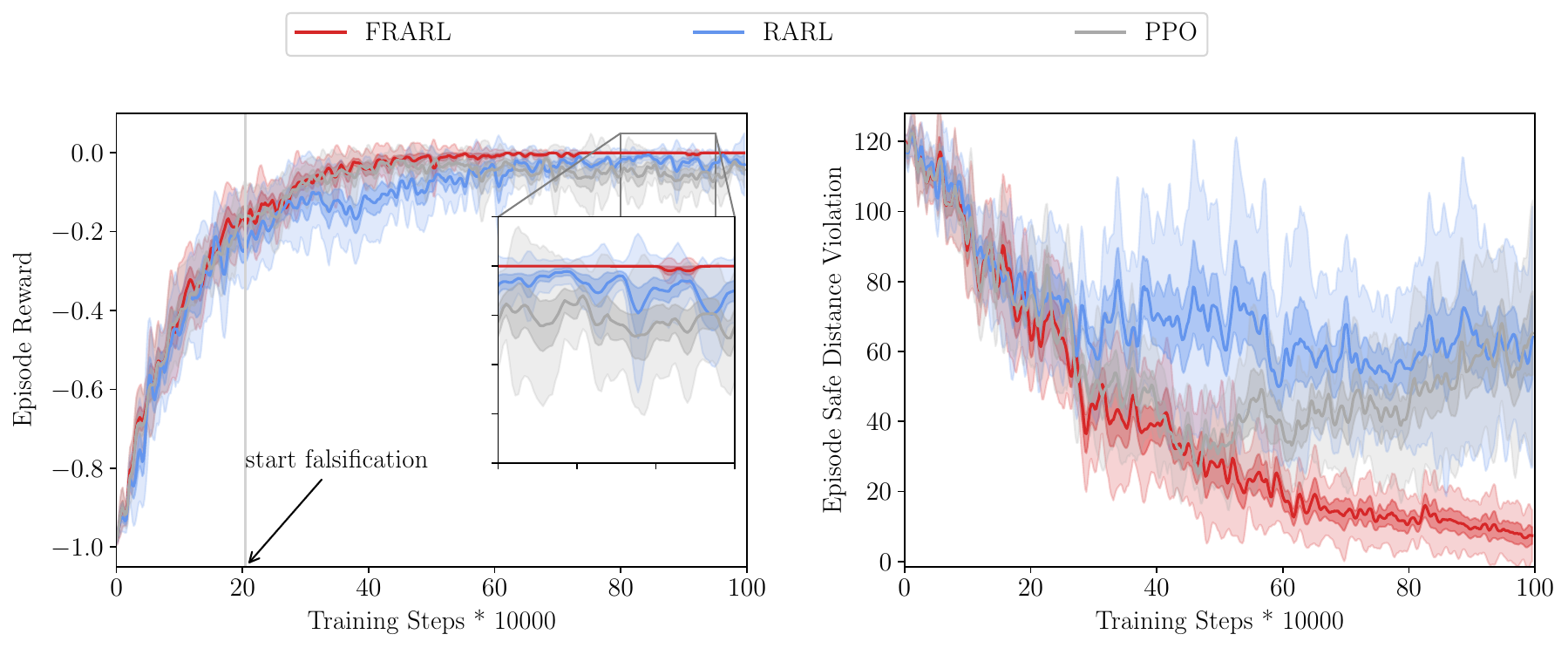}}
	\subfigure[Episode reward and safe distance violation curves on the HighD testing scenarios of the adaptive cruise control system.]{\includegraphics[trim={0 0 0 0.5cm}, clip, scale=0.375]{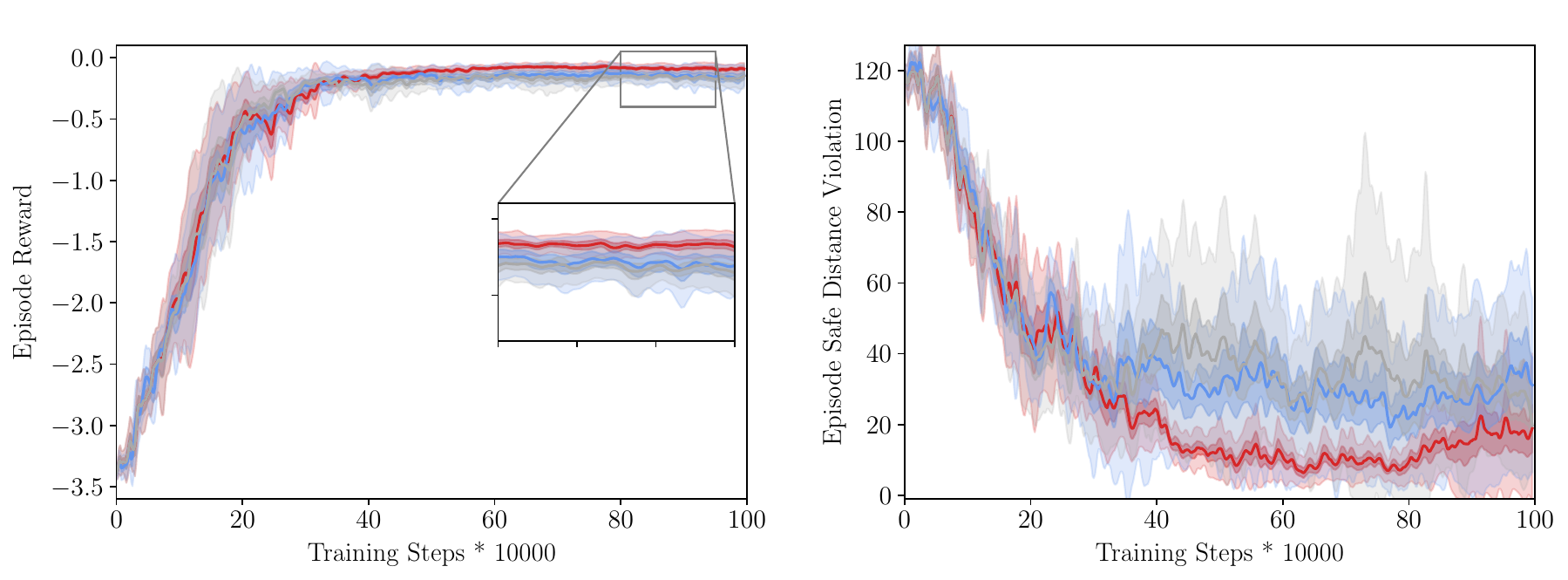}}
	\caption{The learning curves of episode reward and number of safe distance violations on the \textbf{HighD} test scenarios of the BA and ACC system trained using PPO, RARL, and FRARL. For both systems, FRARL showed slightly higher episode reward and much less safe distance violations. In addition, for the BA system, FRARL converged to a zero reward at half of the training steps. Furthermore, FRARL had a lower variance for the episode reward and safe distance violations.}
	\label{fig:learning_curve_highd}
	
\end{figure*}

\begin{figure*}[tb]
	\centering
	\footnotesize
	\subfigure[Episode reward and safe distance violation curves on random testing scenarios of the braking assistant system.]{\includegraphics[trim={0 0 0 0.3cm}, clip, scale=0.375]{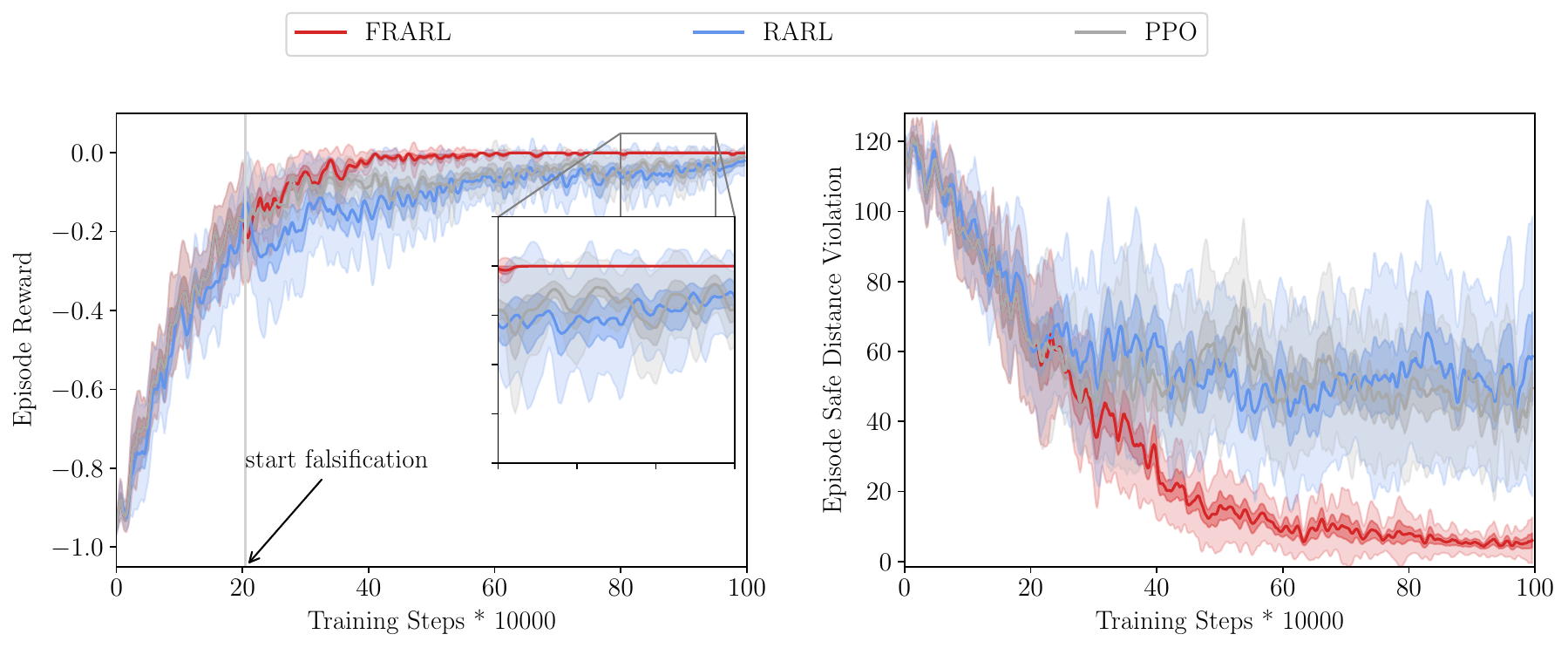}}	
	\subfigure[Episode reward and safe distance violation curves on random testing scenarios of the adaptive cruise control system.]{\includegraphics[trim={0 0 0 0.5cm}, clip, scale=0.375]{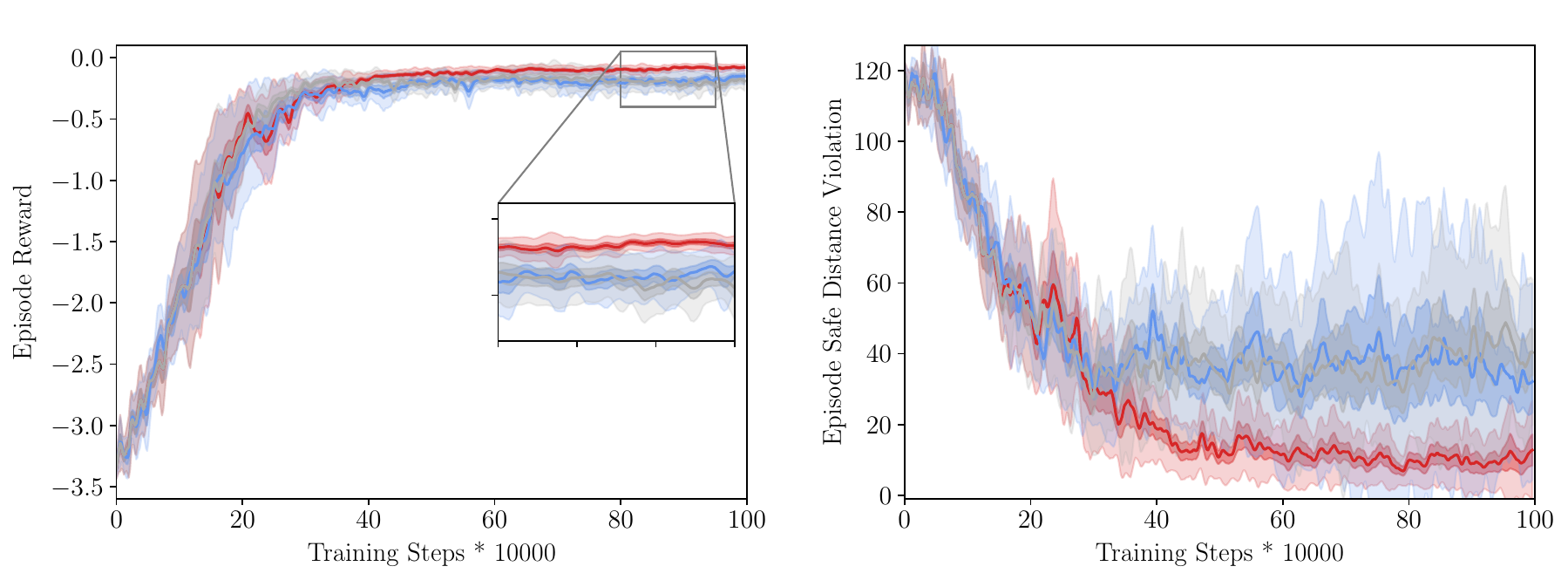}}
	\caption{The learning curves of episode reward and number of safe distance violations on the \textbf{random} test scenarios of the BA and ACC system trained using PPO, RARL, and FRARL. FRARL showed more advantage on random than on HighD scenarios, indicating that training in a falsified environment improves the ability of an RL agent to generalize to unknown scenarios.}
	\label{fig:learning_curve_random}
\end{figure*}

We regard a model as \emph{robust} if it satisfies the safety specification \eqref{eq:MTL_acc} in unseen scenarios, i.e., the HighD and random test scenarios. 
To analyze the safety of the behavior of the agent, we count the number of time steps in which the agent violates the safe distance to the leading vehicle. Figures~\ref{fig:learning_curve_highd} and~\ref{fig:learning_curve_random} depict the episode reward and number of safe distance violations of the BA and ACC systems trained using PPO, RARL, and FRARL in the random and HighD test scenarios, respectively. For both systems, FRARL showed slightly higher episode reward and much less safe distance violations. In addition, for the BA system, FRARL converged to a zero reward at half of the training steps. Furthermore, FRARL had a lower variance for the episode reward and safe distance violations. It also showed more advantage on random than on HighD test scenarios, indicating that training in a falsified environment improves the ability of an RL agent to generalize to unknown scenarios.

\begin{table}[tb]
\caption{Average Rate of Unsafe Behaviors over $28\,037$ HighD test scenarios}
\label{tab:eval_highd}
\vspace{-0.4cm}\vbox{
\begin{center}
\begin{tabular}{l | l l | l l }
\toprule[0.5pt]
                   & \multicolumn{2}{c}{BA}            & \multicolumn{2}{c}{ACC} \\
Violation    & Reverse    & Collision            & Reverse        & Collision        \\
\midrule[0.25pt]
PPO                & 0.34\%     &  4.59\%              &  0.005\%        &   0.24\%          \\
RARL               & 0.009\%    &  2.70\%              &  \textbf{0}    &   0.17\%         \\
FRARL              & \textbf{0} &  \textbf{0.015\%}   &  \textbf{0}    & \textbf{0}\\
\bottomrule[0.5pt]
\end{tabular}
\end{center}}
\vspace{-0.2cm}
\end{table}
%
\begin{table}[tb]
\caption{Average Rate of Unsafe Behaviors over $28\,037$ random test scenarios}
\label{tab:eval_random}
\vspace{-0.4cm}\vbox{
\begin{center}
\begin{tabular}{l | l l | l l }
\toprule[0.5pt]
                   & \multicolumn{2}{c}{BA}            & \multicolumn{2}{c}{ACC} \\
Violation    & Reverse    & Collision            & Reverse        & Collision        \\
\midrule[0.25pt]
PPO                & 0.40\%     &  6.05\%              &  0.028\%        &   0.33\%          \\
RARL               & 0.026\%    &  3.59\%              &  0.013\%    &   0.26\%         \\
FRARL              & \textbf{0.0018\%} &  \textbf{0.025\%}   &  \textbf{0}    & \textbf{0}\\
\bottomrule[0.5pt]
\end{tabular}
\end{center}}
\vspace{-0.4cm}
\end{table}

To further address the robustness of the trained models, we evaluate all models on \SI{28037}{} HighD and random test scenarios and show the average rate of reverse driving and collisions in Tabs.~\ref{tab:eval_highd} and ~\ref{tab:eval_random}, respectively. 
For both systems on both test scenarios, FRARL achieved the lowest rate of collisions and reverse driving. 
FRARL outperformed RARL because an adversarial RL agent seeks scenarios in which the reward of the policy stays low but not necessarily safety-critical scenarios. Thus, after a policy converges to a good behavior, a further increase in safety becomes much more difficult for RARL. Our falsification method instead optimizes the scenarios until safety-critical scenarios are obtained. Therefore, the policies trained in safety-critical scenarios behave much safer than those classically trained.

\section{CONCLUSIONS}
\label{sec:conclusions}

We presented a framework for combining RL with safety falsification methods, which served as an adversarial RL, to improve the robustness of trained policies. By formulating safety requirements in metric temporal logics (MTLs), we spared ourselves the trouble of handcrafting a reward function for the adversary. For BA and ACC systems, we demonstrated that the policies trained using our approach satisfy the safety specification much better in test scenarios and thus are more robust. In the future, we will extend our experiments to more complex driving scenarios, such as urban scenarios. Further, we will integrate traffic rules in our MTL specifications.


\section*{ACKNOWLEDGMENT}

The authors gratefully acknowledge the financial support of this work by the German Research Foundation (DFG) under grant number AL 1185/3-2.

 



%



\balance

\bibliographystyle{IEEEtran}
\bibliography{library}

\end{document}